\documentclass[runningheads]{llncs}

\usepackage[year=2026]{eccv}

\usepackage{eccvabbrv}

\usepackage{graphicx}
\usepackage{booktabs}

\usepackage{amsmath}
\usepackage{amssymb}
\usepackage{color, colortbl}
\usepackage{multirow}
\usepackage{wrapfig}

\usepackage[accsupp]{axessibility}  

\usepackage{hyperref}

\usepackage{orcidlink}

\begin{document}

\title{Device-Conditioned Neural Architecture Search for Efficient Robotic Manipulation}

\titlerunning{Device-Conditioned NAS for Efficient Robotic Manipulation}

\author{
Yiming Wu\inst{1} \and
Huan Wang\inst{2} \and
Zhenghao Chen\inst{3} \and
Ge Yuan\inst{1} \and
Dong Xu\inst{1}
}

\authorrunning{Wu. et al.}

\institute{The University of Hong Kong, China \and
Westlake University, China \and
The University of Newcastle, Australia \\
\email{\{yimingwu, dongxu\}@hku.hk \hspace{12pt} gavinyuan@connect.hku.hk \hspace{12pt} wanghuan@westlake.edu.cn \hspace{12pt} zhenghao.chen@newcastle.edu.au}\\
}

\maketitle

\begin{abstract}
The growing complexity of visuomotor policies poses significant challenges for deployment with heterogeneous robotic hardware constraints. 
However, most existing model-efficient approaches for robotic manipulation are device- and model-specific, lack generalizability, and require time-consuming per-device optimization during the adaptation process.
In this work, we propose a unified framework named \textbf{D}evice-\textbf{C}onditioned \textbf{Q}uantization-\textbf{F}or-\textbf{A}ll (DC-QFA) which amortizes deployment effort with the device-conditioned quantization-aware training and hardware-constrained architecture search. 
Specifically, we introduce a single supernet that spans a rich design space over network architectures and mixed-precision bit-widths. It is optimized with latency- and memory-aware regularization, guided by per-device lookup tables.
With this supernet, for each target platform, we can perform a once-for-all lightweight search to select an optimal subnet without any per-device re-optimization, which enables more generalizable deployment across heterogeneous hardware, and substantially reduces deployment time. To improve long-horizon stability under low precision, we further introduce multi-step on-policy distillation to mitigate error accumulation during closed-loop execution. Extensive experiments on three representative policy backbones, such as DiffusionPolicy-T, MDT-V, and OpenVLA-OFT, demonstrate that our DC-QFA achieves $2\text{-}3\times$ acceleration on edge devices, consumer-grade GPUs, and cloud platforms, with negligible performance drop in task success. Real-world evaluations on an Inovo robot equipped with a force/torque sensor further validates that our low-bit DC-QFA policies maintain stable, contact-rich manipulation even under severe quantization.
\end{abstract}
\section{Introduction}
\label{sec:intro}
Transformer-based visuomotor policies~\cite{chi2025diffusion,brohan2023rt} have significantly advanced robotic manipulation by enabling rich multimodal reasoning and long-horizon planning. However, these models typically incur substantial computational and memory costs, making them difficult to deploy on edge devices such as embedded GPUs or NPUs that are commonly used in real-world robots. In practice, robotic systems often operate under strict latency and memory constraints, creating a gap between high-capacity policies trained on powerful servers and the limited resources available on deployment hardware.

To address this issue, recent studies have explored model compression techniques such as pruning and quantization. Prior studies, including LightDP~\cite{wu2025device} and SQIL~\cite{park2025saliency}, directly compress policies in domain-specific settings to improve inference efficiency. However, these approaches usually require retraining or fine-tuning for each deployment target, making cross-device adaptation expensive. In addition, because they optimize a fixed compressed model for each target separately, the resulting policies can still be suboptimal in the accuracy-efficiency trade-off under heterogeneous hardware budgets.

To overcome these challenges, we explore a \textbf{quantization-for-all (QFA)} approach for embodied visuomotor policies. The goal of QFA is to train a single quantization-aware supernet that supports diverse architectural and precision configurations, from which deployment-ready policies can be obtained via lightweight search without per-device retraining. Building on this formulation, we propose \textbf{Device-Conditioned Quantization-for-All (DC-QFA)}, a unified framework that enables efficient deployment of transformer-based policies across heterogeneous robotic hardware platforms. 
We train a device-conditioned quantization supernet that jointly models architectural and precision configurations under hardware constraints. 
During the training stage, hardware constraints are incorporated using real-device latency and memory profiles. At the deployment (inference) stage, we perform hardware-aware multi-objective search using NSGA-II to identify subnetworks that satisfy device-specific resource budgets while maintaining strong task performance.

While DC-QFA removes the substantial retraining overhead required by previous per-device training methods, it may still suffer a moderate performance gap in long-horizon execution. This is because DC-QFA can introduce slightly larger quantization errors at each time step, which accumulate over time and ultimately lead to unstable behavior in long-horizon manipulation tasks, especially under low-precision computation.
To address this issue, we introduce a \textbf{multi-step on-policy distillation} strategy that aligns the quantized student policy with a full-precision teacher along student rollouts.
Specifically, instead of distilling only one-step predictions, we distill the quantized student along multi-step student-generated trajectories, where a full-precision teacher provides supervision at each visited state and the training horizon is gradually expanded over time.
This mechanism mitigates quantization error accumulation and improves the stability of sequential decision-making policies.

Overall, extensive experiments on multiple embodied manipulation benchmarks and real-robot platforms demonstrate the effectiveness of our framework. In simulation, we conduct cross-benchmark experiments on Push-T, CALVIN, and LIBERO using Diffusion Policy, MDT-V, and OpenVLA-OFT models. The results show that DC-QFA improves the latency-performance trade-off across tasks and architectures. Furthermore, experiments on the long-horizon CALVIN benchmark show that on-policy distillation significantly reduces failure rates in long trajectories. On real robots, we validate the deployment benefits of DC-QFA by training policies with real demonstrations and deploying them directly on edge devices. In summary, our contributions are as follows:
\begin{itemize}
\item We formulate the problem of quantization-for-all (QFA) for embodied visuomotor policies, enabling a single quantization-aware supernet to support deployment across heterogeneous robotic hardware platforms.
\item We propose DC-QFA, a device-conditioned framework that integrates real-device latency and memory constraints into supernet optimization and performs hardware-aware subnet search using NSGA-II algorithm.
\item We further introduce multi-step on-policy distillation to reduce quantization-induced error accumulation and improve long-horizon control consistency compared to single-step baselines.
\item We evaluate DC-QFA across multiple benchmarks and real-robot deployments, achieving $2$--$3\times$ acceleration with negligible performance degradation.
\end{itemize}

\section{Related Work}
\label{sec:related}

\subsection{Visuomotor Policies for Robotic Manipulation}

Imitation learning (IL) has become a dominant paradigm for robotic manipulation by enabling policies to be learned directly from expert demonstrations without requiring explicit reward engineering. Diffusion Policy~\cite{chi2025diffusion} introduces diffusion models for visuomotor trajectory generation and achieves strong performance in high-dimensional manipulation tasks. Subsequent works such as Multimodal Diffusion Transformer (MDT)~\cite{reuss2024multimodal} and Reactive Diffusion Policy~\cite{xue2025reactive} extend this paradigm by incorporating multimodal goal conditioning and reactive feedback.
Vision-Language-Action (VLA) models further extend imitation learning by enabling robots to interpret natural language instructions alongside visual observations. Representative approaches include OpenVLA~\cite{kim2025openvla}, OpenVLA-OFT~\cite{kim2025fine}, $\pi0$~\cite{black2024pi_0}, and $\pi0.5$~\cite{intelligence2025pi_05}. These models combine powerful vision encoders such as DINOv2~\cite{oquab2024dinov2} and SigLIP~\cite{zhai2023sigmoid} with large language models like LLaMA-2~\cite{touvron2023llama} to generate structured action sequences conditioned on multimodal inputs. Large-scale robot learning systems such as SayCan~\cite{brohan2023can}, the RT series~\cite{brohan2023rt,zitkovich2023rt,o2024open,belkhale2024rt}, and the GR series~\cite{wu2023unleashing,cheang2024gr,cheang2025gr} further demonstrate that scaling model capacity and training data significantly improves policy generalization.

Despite their strong performance, these transformer-based visuomotor policies contain hundreds of millions of parameters, posing substantial challenges for real-time deployment on resource-constrained robotic hardware. Our work addresses this challenge by enabling efficient deployment of large visuomotor policies through quantization-aware training and hardware-aware optimization.

\subsection{Efficient Deployment for Robotic Policies}

To reduce inference cost and enable deployment on edge devices, various model compression techniques such as pruning, knowledge distillation, and quantization have been explored. In the robotics domain, LightDP~\cite{wu2025device} and SaliencyQIL~\cite{park2025saliency} propose policy-specific acceleration strategies that combine pruning and quantization to improve inference efficiency. 

Quantization has become one of the most widely used approaches for reducing model size and computational cost. Post-training quantization (PTQ) methods such as SmoothQuant~\cite{xiao2023smoothquant} and AWQ~\cite{lin2024awq} adjust activation and weight distributions to mitigate quantization errors without retraining, while approaches such as QuaRot~\cite{quarot} and ParetoQ~\cite{liu2025paretoq} further improve stability under extremely low-bit inference. Alternatively, quantization-aware training (QAT) simulates quantization effects during training, allowing models to directly optimize performance under low-precision constraints. Hardware-aware QAT methods such as HAQ~\cite{wang2019haq} incorporate latency and energy models to guide precision assignment.

Beyond model compression, recent works also explore system-level strategies to improve the real-time responsiveness of large visuomotor policies. SmolVLA~\cite{shukor2025smolvla} introduces asynchronous policy execution to hide inference latency during action generation. Black \etal~\cite{black2025realtime} propose Real-Time Chunking (RTC), which predicts future action segments while executing the current action block, and later extend it with training-time action conditioning to reduce runtime overhead~\cite{black2025training}. In addition, the SAIL framework~\cite{arachchige2025sail} enables faster-than-demonstration policy execution through error-adaptive guidance and adaptive speed scheduling.

While these approaches improve the runtime responsiveness of large robot policies, they mainly focus on execution scheduling or optimizing a single compressed model. When deploying policies across heterogeneous robotic hardware platforms, existing methods often require repeated retraining or fine-tuning. In contrast, our work enables efficient cross-device deployment through a single quantization-aware training process under the quantization-for-all framework.

\subsection{Once-for-All Quantization}

Once-for-All (OFA) frameworks aim to address the inefficiency of per-device optimization by training a shared supernet that supports multiple architectural configurations. OFA~\cite{cai2019once} introduces a progressive shrinking strategy that allows subnetworks with different depths, widths, and kernel sizes to be sampled without retraining. Building on this idea, One-QAT~\cite{shen2021once} integrates quantization-aware training with neural architecture search, jointly optimizing architecture design and quantization precision.

Recent works such as BatchQuant~\cite{batchquant}, QuantNAS~\cite{gao2024quantnas}, and LLM-QFA~\cite{yi2025one} further extend this paradigm by incorporating quantizer sharing and search-aware quantization strategies. These approaches enable a single supernet to support multiple precision configurations and allow deployment-specific subnetworks to be obtained through lightweight search.

Nevertheless, existing once-for-all quantization frameworks are primarily developed for vision or language models and rarely consider the deployment requirements of embodied policies. In particular, they do not explicitly account for heterogeneous robotic hardware or the stability challenges introduced by quantization in long-horizon sequential decision-making tasks. In contrast, our proposed DC-QFA framework introduces device-conditioned training with real-device latency and memory constraints, enabling efficient specialization of quantized policies for heterogeneous robotic platforms while maintaining stability in long-horizon manipulation tasks.

\section{Methodology}
\begin{figure*}[t]
    \centering
    \includegraphics[width=\linewidth]{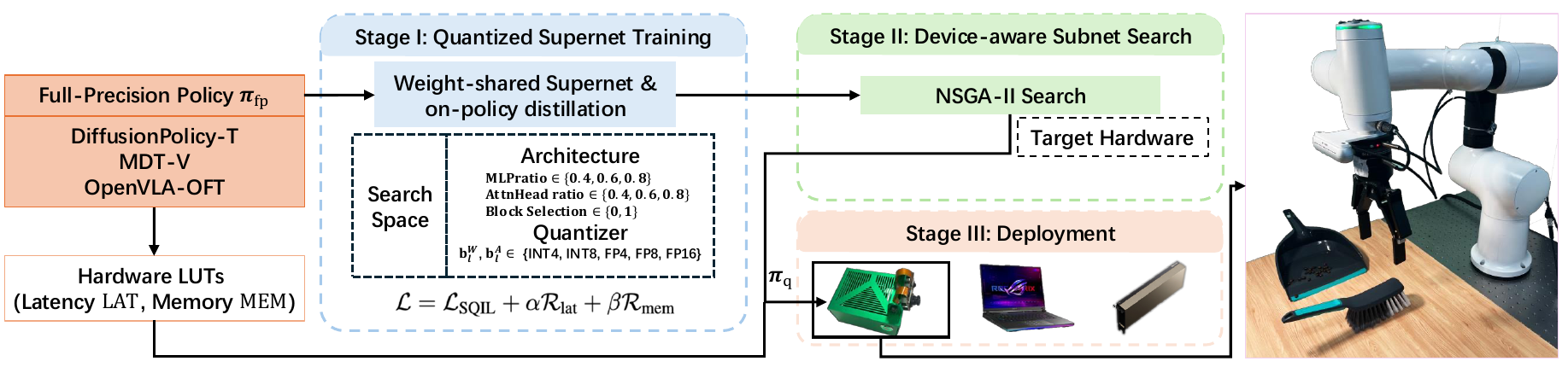}
    \caption{Overview of the proposed DC-QFA pipeline. In \textbf{stage I} (Quantized Supernet Training), we train a weight-sharing supernet with quantization-aware training. In \textbf{stage II} (Device-aware Subnet Search), we search subnets under device-specific latency and memory constraints by optimizing a multi-objective objective guided by measured costs. In \textbf{stage III} (Deployment), we export the selected subnet, calibrate/compile for the target backend, and perform low-bit inference on-device.}
    \label{fig:pipeline}
\end{figure*}

\label{sec:method}
\subsection{Problem Formulation}

We study the problem of deploying large visuomotor policies on heterogeneous robotic hardware platforms with diverse latency and memory constraints. Let $\pi_{\Theta,c}(a_t \mid o_t)$ denote a policy instantiated from a supernet, where $\Theta$ represents the shared supernet parameters and $c \in \mathcal{C}$ denotes a specific subnetwork configuration. Here $o_t$ and $a_t$ represent the observation and action at timestep $t$. Modern transformer-based visuomotor policies typically contain hundreds of millions of parameters, which leads to substantial computational and memory overhead during inference.

To address this challenge, we consider the \textit{quantization-for-all} (QFA) setting. The objective of QFA is to train a single quantization-aware supernet that supports diverse architectural and precision configurations. Deployment-ready subnetworks can then be obtained through lightweight search without retraining for each target device.
A configuration $c = \{(m_l, r_l, h_l, b_l^W, b_l^A)\}_{l=1}^{L}$ specifies architectural and quantization parameters for each layer, where $m_l$ denotes block selection, $r_l$ the MLP expansion ratio, $h_l$ the attention head ratio, and $b_l^W, b_l^A$ the weight and activation bit-widths.

Given a target device $d$ with latency and memory budgets $(B_d^{lat}, B_d^{mem})$, the deployment objective is to identify a configuration that minimizes the policy loss while satisfying hardware constraints:
\begin{equation}
c_d =
\arg\min_{c \in \mathcal{C}}
\mathcal{L}_{policy}(c)
\quad
\text{s.t.}\quad
C_{lat}(d,c) \le B_d^{lat},
\;
C_{mem}(d,c) \le B_d^{mem},
\end{equation}
here $\mathcal{L}_{policy}$ denotes the policy loss used in the supernet training and $C_{lat}, C_{mem}$ represent device-dependent latency and memory costs. Directly solving this constrained optimization for every device is computationally expensive. Our goal is therefore to train a single supernet whose subnetworks remain robust under quantization and can be efficiently specialized for different hardware platforms.

\subsection{Device-Conditioned Supernet Training}

We train the quantization-aware supernet using demonstration data. Let $\mathcal{D} = \{(o_t, a_t)\}$ denote a dataset of observation-action pairs collected from expert demonstrations. During training, we jointly sample a device $d$ from a set of candidate platforms and a configuration $c$ from the supernet search space. The sampled subnet $\pi_{\Theta,c}$ inherits weights from the shared supernet and is optimized using quantization-aware training.

\noindent\textbf{Training objective.} To align the training objective with the deployment objective, we adopt a Lagrangian relaxation of the constrained optimization problem. This results in a hardware-aware training objective of the form:
\begin{equation}
\mathcal{L}_{base}
=
\mathbb{E}_{c \sim \mathcal{C}, (o_t,a_t)\sim \mathcal{D}}
\Big[
\mathcal{L}_{SQIL}(o_t,a_t; c)
+
\alpha \mathcal{R}_{lat}(d,c)
+
\beta \mathcal{R}_{mem}(d,c)
\Big],
\end{equation}
here we adopt loss proposed in SQIL~\cite{park2025saliency} for policy loss calculation.

\noindent\textbf{Hardware-aware regularization.} To incorporate hardware constraints during training, we introduce latency and memory regularization terms:
\begin{align}
\mathcal{R}_{lat}(d,c) &= 
\mathrm{softplus}
\left(
\frac{C_{lat}(d,c)-B_d^{lat}}{B_d^{lat}}
\right), \\
\mathcal{R}_{mem}(d,c) &= 
\mathrm{softplus}
\left(
\frac{C_{mem}(d,c)-B_d^{mem}}{B_d^{mem}}
\right).
\end{align}

Here $C_{lat}(d,c)$ and $C_{mem}(d,c)$ denote estimated latency and memory consumption for configuration $c$ on device $d$. These costs are estimated using lookup tables obtained from real-device profiling. Specifically, each subnet is decomposed into reusable building blocks under different quantization settings. The latency of a subnet is computed by aggregating block-level measurements, while memory usage is estimated from quantized weight and activation statistics.

By sampling device budgets during training, the supernet is repeatedly optimized under different latency-memory trade-offs rather than a single fixed deployment target. As a result, different subnet configurations become specialized for different resource constraints during training.

\subsection{Multi-Step On-Policy Distillation}

Although the base training objective improves quantization robustness, aggressive low-bit inference may still introduce compounding errors during long-horizon execution. To mitigate this effect, we introduce a multi-step on-policy distillation (OPD) objective that aligns the quantized policy with a full-precision teacher along student-generated trajectories.

Let $\pi_T$ denote the full-precision teacher policy instantiated from the largest configuration of the supernet. The quantized student policy $\pi_{\Theta,c}$ is rolled out for $K$ steps to produce a trajectory
\begin{equation}
\tau = \{(o_1,a_1), (o_2,a_2), \dots, (o_K,a_K)\}.
\end{equation}

At each step, the teacher provides supervision for the student action distribution. The distillation loss is defined as
\begin{equation}
\mathcal{L}_{OPD}
=
\frac{1}{K}
\sum_{t=1}^{K}
w_t
\,\mathit{D}
\big(
\pi_T(\cdot|o_t),
\pi_{\Theta,c}(\cdot|o_t)
\big),
\end{equation}
where $\mathit{D}$ denotes a distribution matching operator. In practice, $\mathit{D}$ corresponds to mean squared error for continuous actions. The final training objective combines the base training loss with the OPD loss:
\begin{equation}
\mathcal{L}_{total}
=
\mathcal{L}_{base}
+
\gamma \mathcal{L}_{OPD}.
\end{equation}

During training, we progressively increase the rollout horizon $K$, transitioning from teacher-aligned supervision to student-driven rollouts. This strategy improves the stability of quantized policies by explicitly reducing error accumulation along long-horizon trajectories.
\section{Experiments}
\label{sec:experiments}
\subsection{Experimental Setup}
\label{sec:experimental_setup}
\begin{figure}[tp]
    \centering
    \includegraphics[width=0.95\linewidth]{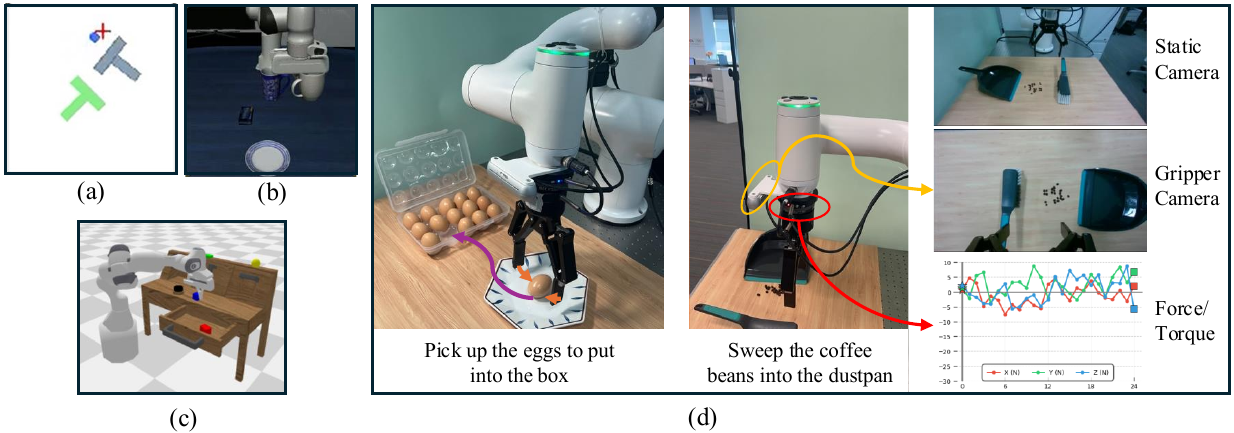}
    \caption{Experimental setups in simulation and the real world. (a) Push-T. (b) LIBERO. (c) CALVIN. (d) Real robot tasks, including object handling (egg) and tool use (brush).}
    \label{fig:experimental_setup}
\end{figure}

We evaluate DC-QFA on three widely-used robotic manipulation benchmarks:
\begin{itemize}
\item \textbf{Push-T}~\cite{chi2025diffusion}, a planar pushing task that requires manipulating a T-shaped block into a goal region. The evaluation metric is task success rate.
\item \textbf{CALVIN}~\cite{mees2022calvin}, a long-horizon language-conditioned manipulation benchmark. Following prior work, we report the average rollout length on the \texttt{D$\rightarrow$D} and \texttt{ABC$\rightarrow$D} splits.
\item \textbf{LIBERO}~\cite{liu2023libero}, a multi-task manipulation benchmark containing 130 tasks across four suites: LIBERO-Spatial, LIBERO-Object, LIBERO-Goal, and LIBERO-Long. We report the average success rate across the four suites.
\end{itemize}
We evaluate our proposed DC-QFA framework on three representative manipulation benchmarks (\ie, DiffusionPolicy-T~\cite{chi2025diffusion}, MDT-V~\cite{reuss2024multimodal}, and OpenVLA-OFT~\cite{kim2025fine}) across diverse manipulation benchmarks. These models span distinct network designs (\eg, ResNet-based visual encoders, Perceiver modules, and Transformer-based diffusion policies) and vary in their input modalities, policy lengths, and diffusion rollout structures.

All experiments are conducted in simulation with consistent evaluation scripts; we additionally report real-device deployment results to assess latency and quantization robustness. Beyond simulation, we validate the proposed method in a real-world setting on an Inovo robot arm equipped with a Robotiq FT 300-S force/torque sensor. The sensor provides high-frequency wrist force feedback during contact-rich manipulation (\eg, sweeping beans and handling eggs). The complete experimental setup is illustrated in \cref{fig:experimental_setup}.

\subsection{Baselines and Metrics}
We compare our DC-QFA against multiple baselines from quantization and NAS literature: \textbf{PTQ-based methods} (\ie, SmoothQuant~\cite{xiao2023smoothquant} and AWQ~\cite{lin2024awq} with calibration dataset size of 256), \textbf{QAT-based methods} (\ie, SQIL~\cite{park2025saliency}, which applies quantization-aware imitation learning to diffusion policies), and \textbf{QFA-based methods} (\ie, BatchQuant~\cite{batchquant}, a method integrating quantizer robustness into NAS).
Following prior work, we report results under W8A8 and W4A4 quantization settings for fair comparison unless otherwise specified.

\subsection{Experiments on Simulation Benchmarks}

\subsubsection{Experiments on OpenVLA-OFT}
\label{sec:openval-oft}
\begin{table}
\caption{Performance of OpenVLA-OFT~\cite{kim2025fine} on the LIBERO~\cite{liu2023libero} benchmark. We compare FP16 and quantized variants at FP8/W8A8 and W4A4 using SmoothQuant~\cite{xiao2023smoothquant}, AWQ~\cite{lin2024awq}, SQIL~\cite{park2025saliency}, and BatchQuant~\cite{batchquant}. Metrics are success rates (\%) on four subsets; Avg is the mean (higher is better). DC-QFA (ours) is highlighted and matches or surpasses SQIL at the same bit-width.}
\vspace{-1em}
  \centering
  \resizebox{0.75\columnwidth}{!}{
  \begin{tabular}{lccccccc}
  \toprule
  \multirow{2}{*}{Method} & \multirow{2}{*}{Bit-width} &  \multirow{2}{*}{Model Size}  & \multicolumn{5}{c}{LIBERO} \\ \cline{4-8}
  &  &  & Spatial & Object & Goal & Long & Avg$\uparrow$ \\
  \hline
  OpenVLA-OFT    & FP16                  & 15.1 GB                        & 97.6 & 98.4 & 97.9 & 94.5 & 97.1 \\
  \hline
  w/ BatchQuant  & FP8                   & 7.5 GB                     & 96.8 & 98.2 & 96.2 & 93.2 & 96.1 \\
  \rowcolor[HTML]{F2F2F2}{DC-QFA (Ours)}  &                       &                            & {97.2} & {98.0}   & {97.2} & {93.9} & \textbf{96.6} \\
  \hline
  w/ SmoothQuant & \multirow{5}{*}{W8A8} & \multirow{5}{*}{7.5 GB} & 94.8 & 93.1 & 93.4 & 90.7 & 93 \\
  w/ AWQ         &                       &                         & 94.7 & 93.5 & 93.8 & 91.1 & 93.3 \\
  w/ SQIL        &                       &                         & 97.0 & 98.1 & 95.9 & 93.2 & \textbf{96.1} \\
  w/ BatchQuant  &                       &                         & 96.7 & 97.8 & 95.1 & 92.1 & 95.4 \\
  \rowcolor[HTML]{F2F2F2}{DC-QFA (Ours)}  &                       &                         & 96.9 & 97.5  &  95.7  & 93.7 & {96.0} \\
  \hline
  w/ SmoothQuant & \multirow{5}{*}{W4A4} & \multirow{5}{*}{3.8 GB} & 94.2 & 92.9 & 93 & 90.0 & 92.6 \\
  w/ AWQ         &                       &                         & 94.9 & 93.6 & 93.8 & 90.7 & 93.0 \\
  w/ SQIL        &                       &                         & 96.8 & 97.6 & 95.4 & 93.0 & 95.8 \\
  w/ BatchQuant  &                       &                         & 96.8 & 97.6 & 94.8 & 92.5 & 95.4 \\
  \rowcolor[HTML]{F2F2F2}{DC-QFA (Ours)}  &                       &                         & {96.5} & {97.8} & {96.1} & {93.1} & \textbf{95.9} \\
  \bottomrule
  \end{tabular}
}
\vspace{-2em}
\label{tab:openvla-oft}
\end{table}

\cref{tab:openvla-oft} reports the performance of OpenVLA-OFT on the LIBERO benchmark under different quantization settings. 
The full-precision FP16 model achieves an average success rate of 97.1\% across the four suites (Spatial, Object, Goal, and Long) with a model size of 15.1GB. 

Under FP8 quantization, BatchQuant reduces the model size by $2\times$ to 7.5GB with an average success rate of 96.1\%, while DC-QFA recovers most of the performance with 96.6\% average success, remaining within $0.7\%$ of the FP16 baseline across all subsets.
In the W8A8 setting, PTQ methods such as SmoothQuant and AWQ show noticeable degradation (around 93\% average success), whereas QAT-based methods SQIL and BatchQuant significantly improve performance to 96.1\% and 95.4\%, respectively. DC-QFA achieves 96.0\% average success, close to the best QAT baseline while maintaining balanced performance across all four task suites.
Under the more aggressive W4A4 configuration (3.8GB, $\approx4\times$ compression), PTQ methods further drop to around 92.6\%--93.0\% average success. In contrast, QAT-based methods maintain stronger performance. DC-QFA achieves the best overall result with 95.9\% average success, slightly surpassing SQIL and BatchQuant while remaining close to the FP16 baseline despite substantial compression. These results demonstrate that DC-QFA effectively preserves the generalization capability of OpenVLA-OFT under aggressive quantization while enabling significantly more efficient deployment.

\subsubsection{Experiments on DiffusionPolicy Transformer}
\label{sec:dpt}
\begin{wraptable}{r}{0.5\columnwidth}
\vspace{-3em}
\caption{
Results of DiffusionPolicy-T~\cite{chi2025diffusion} on the Push-T dataset, where we compare various quantization techniques and their impact on model size and success rate.
}
  \centering
  \resizebox{\linewidth}{!}{
  \begin{tabular}{lccc}
  \toprule
  Method & Bit-width & Model Size & Success Rate $\uparrow$ \\
  \hline
  DiffusionPolicy-T & FP16 & 38.48 MB & 0.772$\pm$0.039 \\
  \hline

  {w/ BatchQuant}                     & FP8                   & 19.24 MB              & 0.759$\pm$0.021 \\
  \rowcolor[HTML]{F2F2F2}{DC-QFA (Ours)}   &                       &                           & \textbf{0.770$\pm$0.016} \\ \hline

  w/ SmoothQuant & \multirow{5}{*}{W8A8} & \multirow{5}{*}{19.24 MB} & 0.729$\pm$0.011 \\
  w/ AWQ          &                      &                           & 0.733$\pm$0.022 \\
  w/ SQIL         &                      &                           & 0.766$\pm$0.019 \\
  w/ BatchQuant   &                      &                           & 0.757$\pm$0.017 \\
  \rowcolor[HTML]{F2F2F2}{DC-QFA (Ours)}   &                      &                         & \textbf{0.769$\pm$0.011} \\ \hline

  w/ SmoothQuant & \multirow{5}{*}{W4A4} & \multirow{5}{*}{9.62 MB} & 0.727$\pm$0.023 \\
  w/ AWQ          &                      &                           & 0.723$\pm$0.012 \\
  w/ SQIL         &                      &                           & 0.768$\pm$0.013 \\
  w/ BatchQuant   &                      &                           & 0.761$\pm$0.016 \\
  \rowcolor[HTML]{F2F2F2}{DC-QFA (Ours)}   &                      &                         & \textbf{0.770$\pm$0.013} \\
  \bottomrule
  \end{tabular}
}
\vspace{-2em}

\label{tab:dpt}

\end{wraptable}
DiffusionPolicy-T is evaluated on \textbf{Push-T}~\cite{chi2025diffusion}, a planar manipulation task that requires pushing a T-shaped block into a goal region. The initial poses of the block and end-effector are randomized, and the policy receives RGB observations together with keypoint-based states. Following the evaluation protocol of DiffusionPolicy~\cite{chi2025diffusion}, we report the task success rate averaged over multiple rollouts. Training DC-QFA for DiffusionPolicy-T requires approximately 85 GPU hours.

\cref{tab:dpt} reports the quantitative comparison under different quantization settings. The FP16 baseline achieves a success rate of $0.772\pm0.039$ with a model size of 38.48MB. Under FP8 quantization, BatchQuant reduces the model size by $2\times$ to 19.24MB with a success rate of $0.759\pm0.021$, while DC-QFA recovers nearly full-precision performance with $0.770\pm0.016$. 
In the W8A8 setting, PTQ methods such as SmoothQuant and AWQ show noticeable degradation (around $0.73$), whereas QAT-based methods SQIL and BatchQuant improve performance to $0.766$ and $0.757$, respectively. DC-QFA further achieves $0.769\pm0.011$, matching the best QAT baseline while maintaining the same model size. 
Under the more aggressive W4A4 configuration (9.62MB, $\approx4\times$ compression), PTQ methods degrade to around $0.72$ success, while SQIL and BatchQuant retain stronger performance ($0.768$ and $0.761$). DC-QFA achieves $0.770\pm0.013$, outperforming all baselines under the same quantization level and remaining essentially on par with the FP16 model despite the substantial compression. These results demonstrate that DC-QFA effectively preserves the closed-loop behavior of DiffusionPolicy-T under aggressive quantization while enabling compact deployable policies.

\subsubsection{Experiments on Multimodal Diffusion Transformer}
\label{sec:mdt}
\begin{table*}[t]
\centering
\caption{
Results of MDT-V~\cite{reuss2024multimodal} on the CALVIN~\cite{mees2022calvin} and LIBERO~\cite{liu2023libero} datasets, including performance metrics across various quantization methods and bit-widths. The table presents the model size, accuracy, and average performance (average length on CALVIN and success rate on LIBERO) for each method, highlighting the effectiveness of our DC-QFA when compared to other quantization methods.
}
\resizebox{0.95\textwidth}{!}{%

\begin{tabular}{lccccccccccc}
\toprule
\multirow{2}{*}{Method} & \multirow{2}{*}{Bit-width} &  \multirow{2}{*}{Model Size} & \multicolumn{2}{c}{CALVIN}
& \multicolumn{5}{c}{LIBERO} \\ \cline{4-10}
 &  &  & \texttt{D$\rightarrow$D$\uparrow$} & \texttt{ABC$\rightarrow$D$\uparrow$}  & Spatial$\uparrow$ & Object$\uparrow$ & Goal$\uparrow$ & Long$\uparrow$ & Average$\uparrow$ \\
\hline
MDT-V (FP16) & FP16 & 45.04 MB & 4.52$\pm$0.02 & 3.72$\pm$0.06 & 77.9 & 87.5 & 75.4 & 65 & 76.5 \\
\hline


w/ BatchQuant & FP8                  & 22.52 MB                  & 4.45$\pm$0.10 & 3.65$\pm$0.05  & 77.2 & 86.9 & 75.1 & 64.8 & 76.0 \\
\rowcolor[HTML]{F2F2F2}{DC-QFA (Ours)} &                       &                           & \textbf{4.47$\pm$0.07} & \textbf{3.67$\pm$0.09} & {77.4} & {87.4} & {75.5} & {65.1} & \textbf{76.4} \\
\hline

w/ SmoothQuant & \multirow{5}{*}{W8A8} & \multirow{5}{*}{22.52 MB} & 4.23$\pm$0.04 & 3.57$\pm$0.09 & 72.8 & 81.9 & 71.5 & 62.1 & 72.0 \\
w/ AWQ         &                       &                           & 4.21$\pm$0.03 & 3.55$\pm$0.06 & 72.8 & 82.1 & 72.2 & 62.9 & 72.5 \\
w/ SQIL        &                       &                           & 4.43$\pm$0.05 & \textbf{3.67$\pm$0.07} & 76.5 & 87.0 & 74.9 & 64.2 & 75.7 \\
w/ BatchQuant  &                       &                           & 4.42$\pm$0.14 & 3.58$\pm$0.06 & 76.7 & 86.8 & 75.2 & 64.6 & 75.8 \\
\rowcolor[HTML]{F2F2F2}{DC-QFA (Ours)}  &                       &                           & \textbf{4.48$\pm$0.11} & {3.64$\pm$0.07} & {77.2} & {87.2} & {75.1} & {64.6} & \textbf{76.0} \\
\hline

w/ SmoothQuant & \multirow{5}{*}{W4A4} & \multirow{5}{*}{11.26 MB} & 4.19$\pm$0.08 & 3.43$\pm$0.12 & 73.4 & 82.2 & 71.9 & 61.9 & 72.4 \\
w/ AWQ         &                       &                           & 4.22$\pm$0.09 & 3.48$\pm$0.16 & 72.5 & 82.4 & 72.4 & 62.4 & 72.4 \\
w/ SQIL        &                       &                           & 4.39$\pm$0.06 & 3.63$\pm$0.10 & 76.3 & 86.8 & 75.4 & 64.1 & 75.4 \\
w/ BatchQuant  &                       &                           & 4.33$\pm$0.12 & 3.59$\pm$0.08 & 76.6 & 86.2 & 75.1 & 64.3 & {75.6} \\
\rowcolor[HTML]{F2F2F2}{DC-QFA (Ours)}  &                       &                           & \textbf{4.45$\pm$0.09} & \textbf{3.66$\pm$0.13} & {77.3} & {87.0} & {75.2} & {64.1} & \textbf{75.9} \\
\bottomrule
\end{tabular}

}
\label{tab:mdt}
\vspace{-2em}
\end{table*}

Multimodal Diffusion Transformer~\cite{reuss2024multimodal} is a multimodal diffusion-based policy that integrates visual observations and language goals to produce versatile manipulation behaviors. In our experiments, we use the Voltron-based variant MDT-V as a unified backbone and evaluate it on both CALVIN and LIBERO to test whether DC-QFA can compress a multimodal policy without sacrificing generality. Training DC-QFA for MDT-V requires approximately 156 GPU hours.

CALVIN~\cite{mees2022calvin} is a long-horizon manipulation benchmark where agents must execute sequences of language-conditioned skills. Following prior work, we report the average rollout length on the \texttt{D$\rightarrow$D} and \texttt{ABC$\rightarrow$D} splits. LIBERO~\cite{liu2023libero} evaluates generalization across diverse manipulation tasks, and we report success rates on four suites: Spatial, Object, Goal, and Long.

\cref{tab:mdt} summarizes the results. The FP16 MDT-V baseline achieves rollout lengths of $4.52\pm0.02$ on \texttt{D$\rightarrow$D} and $3.72\pm0.06$ on \texttt{ABC$\rightarrow$D}, with a LIBERO average success rate of 76.5. Under FP8 quantization, BatchQuant reduces the model size by $2\times$ to 22.52MB with rollout lengths of $4.45\pm0.10$ and $3.65\pm0.05$, while DC-QFA slightly improves these results to $4.47\pm0.07$ and $3.67\pm0.09$, achieving a LIBERO average of 76.4 and remaining close to the FP16 baseline.

In the W8A8 setting, PTQ methods such as SmoothQuant and AWQ suffer noticeable degradation on both CALVIN and LIBERO (around 72 average LIBERO success). QAT-based methods substantially recover performance, with SQIL and BatchQuant achieving 75.7 and 75.8 average success on LIBERO. DC-QFA remains competitive on CALVIN performance to $4.48\pm0.11$ on \texttt{D$\rightarrow$D} and achieves a LIBERO average of 76.0, essentially matching the FP16 model while remaining quantized.

Under the more aggressive W4A4 configuration (11.26MB, $\approx4\times$ compression), PTQ methods again exhibit larger performance drops, while QAT-based approaches retain stronger results. DC-QFA achieves $4.45\pm0.09$ and $3.66\pm0.13$ on the CALVIN splits and reaches a LIBERO average success rate of 75.9, remaining close to the FP16 baseline despite the significant compression. These results demonstrate that DC-QFA consistently preserves the behavior of MDT-V under aggressive quantization while outperforming PTQ methods and matching or slightly improving over existing QAT baselines across both CALVIN and LIBERO benchmarks.

\subsubsection{Long-Horizon Stability Analysis}
\label{sec:long_horizon}
\begin{wraptable}{r}{0.55\columnwidth}
\vspace{-3em}
\caption{
Long-horizon evaluation on the CALVIN~\cite{mees2022calvin} benchmark using MDT-V~\cite{reuss2024multimodal}. 
We compare the quantized policies trained with DC-QFA and DC-QFA+OPD under FP8, W8A8, and W4A4 settings. 
Metrics report the average rollout length on the \texttt{D$\rightarrow$D} and \texttt{ABC$\rightarrow$D} splits (higher is better). 
While DC-QFA preserves most of the full-precision performance under quantization, incorporating OPD further improves rollout stability and consistently recovers performance closer to the FP16 baseline in long-horizon tasks.
}
  \centering
  \resizebox{\linewidth}{!}{
  \begin{tabular}{lcccc}
  \toprule
  \multirow{2}{*}{Method} & \multirow{2}{*}{Bit-width} &  \multirow{2}{*}{Model Size} & \multicolumn{2}{c}{CALVIN} \\ \cline{4-5}
  &  &  & \texttt{D$\rightarrow$D$\uparrow$} & \texttt{ABC$\rightarrow$D$\uparrow$} \\
  \hline
  MDT-V  & FP16 & 45.04 MB & 4.52$\pm$0.02 & 3.72$\pm$0.06 \\
  \hline

  {DC-QFA} & \multirow{2}{*}{FP8} & \multirow{2}{*}{22.52 MB} & 4.47$\pm$0.07 & 3.67$\pm$0.09 \\
  {DC-QFA + OPD} &                   &                           & \textbf{4.51$\pm$0.04} & \textbf{3.72$\pm$0.07} \\
  \hline

  {DC-QFA}  & \multirow{2}{*}{W8A8} & \multirow{2}{*}{22.52 MB} & 4.48$\pm$0.11 & {3.64$\pm$0.07} \\
  {DC-QFA + OPD} &                       &                           & \textbf{4.50$\pm$0.07} & \textbf{3.69$\pm$0.09} \\
  \hline

  {DC-QFA}  & \multirow{2}{*}{W4A4} & \multirow{2}{*}{11.26 MB} & 4.45$\pm$0.09 & 3.66$\pm$0.13 \\
  {DC-QFA + OPD} &                  &                           & \textbf{4.47$\pm$0.09} & \textbf{3.67$\pm$0.08} \\
  \bottomrule
  \end{tabular}
}
\vspace{-2em}
\label{tab:long_horizon}
\end{wraptable}

We evaluate the effectiveness of on-policy distillation (OPD) in improving the stability of quantized policies in long-horizon tasks. To isolate this effect, we conduct the analysis on the CALVIN benchmark, which requires sequential execution of multiple manipulation skills and therefore exposes errors introduced by low-precision inference.

\cref{tab:long_horizon} compares DC-QFA with and without OPD under different quantization levels. While the DC-QFA baseline already preserves most of the full-precision performance, OPD consistently improves rollout stability across all bit-width settings. Under FP8 quantization, incorporating OPD increases the rollout length from $4.47$ to $4.51$ on \texttt{D$\rightarrow$D} and from $3.67$ to $3.72$ on \texttt{ABC$\rightarrow$D}, effectively recovering the full-precision performance. Similar improvements are observed for W8A8, where OPD raises the results from $4.48$ to $4.50$ and from $3.64$ to $3.69$, respectively. Even under the more aggressive W4A4 setting, OPD still provides consistent gains, increasing the rollout length from $4.45$ to $4.47$ on \texttt{D$\rightarrow$D} and from $3.66$ to $3.67$ on \texttt{ABC$\rightarrow$D}.

These results indicate that OPD effectively mitigates the errors introduced by quantization. By providing trajectory-level supervision through on-policy rollouts, OPD improves the stability of quantized policies during long-horizon execution while maintaining the efficiency advantages of low-bit inference.

\subsection{Experiments on Real-world Setting}
\label{sec:real-world}
\begin{table}[t]
\centering
\caption{
Real-world manipulation results on the Inovo robotic platform under different quantization settings. 
We evaluate four tasks including \textbf{Task 1: sweeping coffee beans into a dustpan}, 
\textbf{Task 2: picking up eggs and placing them into a box}, 
\textbf{Task 3: inserting a red pepper into a cup}, and 
\textbf{Task 4: picking up a carrot and placing it into a bowl}. 
Each task is evaluated over 20 trials. 
We compare FP16 base model, a PTQ baseline using AWQ~\cite{lin2024awq} with W4A4 quantization, W8A8 policies, and our DC-QFA W4A4 policies. 
Results show that DC-QFA significantly improves the robustness of low-bit policies in real-world manipulation tasks.
}
\resizebox{0.99\columnwidth}{!}{%
\begin{tabular}{l c c c c c c}
\toprule
Metric & OpenVLA-OFT~(T1) & OpenVLA-OFT~(T2) & DiffusionPolicy-T~(T3) & DiffusionPolicy-T~(T4) & MDT-V~(T3) & MDT-V~(T4) \\
\midrule
Base Model (FP16) & 18 & 14 & 11 & 13 & 13 & 14 \\
AWQ (W4A4) & 10 & 8 & 5 & 5 & 4 & 6 \\ \hline
DC-QFA (W8A8) & 18 & 14 & 10 & 13 & 11 & 12 \\
DC-QFA (W4A4) & 15 & 13 & 10 & 13 & 11 & 12 \\
\bottomrule
\end{tabular}
}

\label{tab:real-world}
\vspace{-2em}
\end{table}

We evaluate the real-world deployment capability of DC-QFA on an Inovo 6-DOF robotic arm equipped with a parallel gripper and a Robotiq FT 300-S force/torque sensor. The experiments aim to validate whether policies trained with DC-QFA maintain robust manipulation performance under low-bit inference in real physical environments.

The robot executes policies using TensorRT inference on a Jetson Orin NX edge device. Visual observations are captured using an RGBD camera mounted on the gripper and the front view of the robotic arm, while the force/torque sensor attached on the gripper provides contact feedback during manipulation. We evaluate three policy architectures (DiffusionPolicy-T, MDT-V, and OpenVLA-OFT) under multiple quantization settings including FP16, W8A8, and W4A4.
We evaluate four representative manipulation tasks involving contact-rich object interactions: \textbf{Task 1: sweeping beans}—the robot sweeps scattered coffee beans into a dustpan; \textbf{Task 2: egg picking}—the robot grasps eggs and places them into a container; \textbf{Task 3: pepper insertion}—the robot inserts a red pepper into a cup; and \textbf{Task 4: carrot pick-and-place}—the robot picks up a carrot and places it into a bowl. Each task is executed for 20 trials, and the number of successful executions is recorded.

\cref{tab:real-world} reports real-world results on four contact-rich tasks executed on an Inovo robot with TensorRT inference on Jetson Orin NX. We evaluate OpenVLA-OFT on Tasks 1 and 2 under FP16, W8A8, and W4A4, and compare DiffusionPolicy-T and MDT-V against AWQ and FP16 on Tasks 3 and 4.

For OpenVLA-OFT, W8A8 matches the FP16 baseline on both tasks (18/20 and 14/20), while W4A4 still achieves 15/20 and 13/20. On Tasks 3 and 4, DC-QFA improves markedly over AWQ: DiffusionPolicy-T rises from 5/20 and 5/20 to 10/20 and 13/20, and MDT-V rises from 4/20 and 6/20 to 11/20 and 12/20. The remaining gap to FP16 is small relative to the gain over PTQ.

\textbf{Discussion.}
Across all real-world tasks, DC-QFA consistently preserves manipulation performance under aggressive quantization. Even under W4A4 inference, policies trained with DC-QFA maintain stable closed-loop control and substantially outperform PTQ baselines. These results demonstrate that DC-QFA enables reliable deployment of large visuomotor policies on edge hardware without sacrificing task success.

\cref{tab:real-world} summarizes real-world task success rates under different quantization bit-widths on the Inovo robot platform. We evaluate two everyday manipulation tasks: (1) sweeping coffee beans into a dustpan and (2) picking up eggs and placing them into a box. Each task is run for 20 trials. We report the number of successful trials for FP16, W8A8, and W4A4 policies from the OpenVLA-OFT. The robot is equipped with a wrist-mounted Robotiq FT 300-S force/torque sensor, which provides contact feedback during execution. Deployment uses TensorRT on a Jetson Orin NX. We sample a 24-layer Transformer subset of OpenVLA-OFT, yielding a 7.9GB FP16 runtime memory footprint.

For the two OpenVLA-OFT tasks, DC-QFA with W8A8 exactly matches the FP16 baseline, achieving 18/20 successes on Task~1 and 14/20 on Task~2, while DC-QFA with W4A4 still attains 15/20 and 13/20 successes, respectively. Compared with the PTQ baseline AWQ under W4A4, which drops to 10/20 and 8/20, DC-QFA recovers most of the lost performance even at 4-bit precision. A similar trend is observed on Tasks~3 and~4: for DiffusionPolicy-T, AWQ achieves only 5/20 and 5/20 successes, whereas DC-QFA reaches 10/20 and 13/20, remaining close to the FP16 baseline of 11/20 and 13/20; for MDT-V, AWQ obtains 4/20 and 6/20, while DC-QFA improves them to 11/20 and 12/20, compared with 13/20 and 14/20 for FP16. Overall, these results show that DC-QFA consistently preserves real-world manipulation robustness under low-bit deployment and substantially outperforms standard PTQ, especially in the challenging W4A4 regime.

\subsection{Pareto Frontier and Mixed-Precision Analysis}
\label{sec:pareto_frontier}

\begin{wrapfigure}{r}{0.4\textwidth}
    \vspace{-2em}
    \centering
    \includegraphics[width=\linewidth]{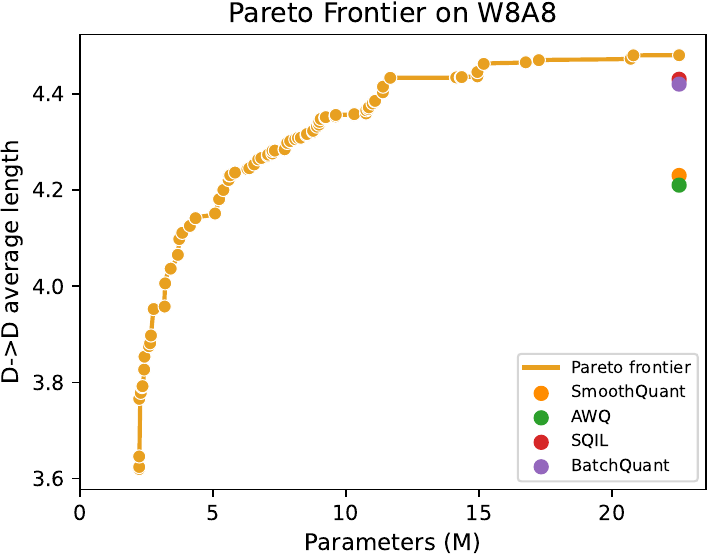}
    \vspace{-2.5em}
    \caption{Pareto curves of performance-parameter trade-offs based on the MDT-V W8A8 setting.}
    \label{fig:pareto}
    \vspace{-3em}
\end{wrapfigure}
\cref{fig:pareto} visualizes the W8A8 Pareto frontier of MDT-V on CALVIN \texttt{D$\rightarrow$D} in the parameter-performance plane. Each blue dot corresponds to a quantized subnet sampled by our DC-QFA, with the $x$-axis showing the number of parameters (in millions) and the $y$-axis showing the average rollout length. 
As model size increases from $\sim$2.5M to $\sim$7M parameters, performance improves from around 3.6 to above 4.2, after which the curve gradually saturates and flattens between 7M and 22M parameters at around 4.4. 
The baseline methods all lie near the right edge of the plot with $\approx$22M parameters, but are either strictly below or very close to our DC-QFA frontier: for similar parameter counts, our DC-QFA subnets attain equal or higher average rollout length. For a fixed performance level, DC-QFA offers substantially smaller models. This indicates that DC-QFA not only preserves the performance of strong QAT baselines such as SQIL, but also identifies more parameter-efficient subnets that consistently outperform PTQ/QAT baselines in the accuracy-model size trade-off.

\begin{wrapfigure}{r}{0.45\textwidth}
\vspace{-2em}
    \centering
    \includegraphics[width=0.95\linewidth]{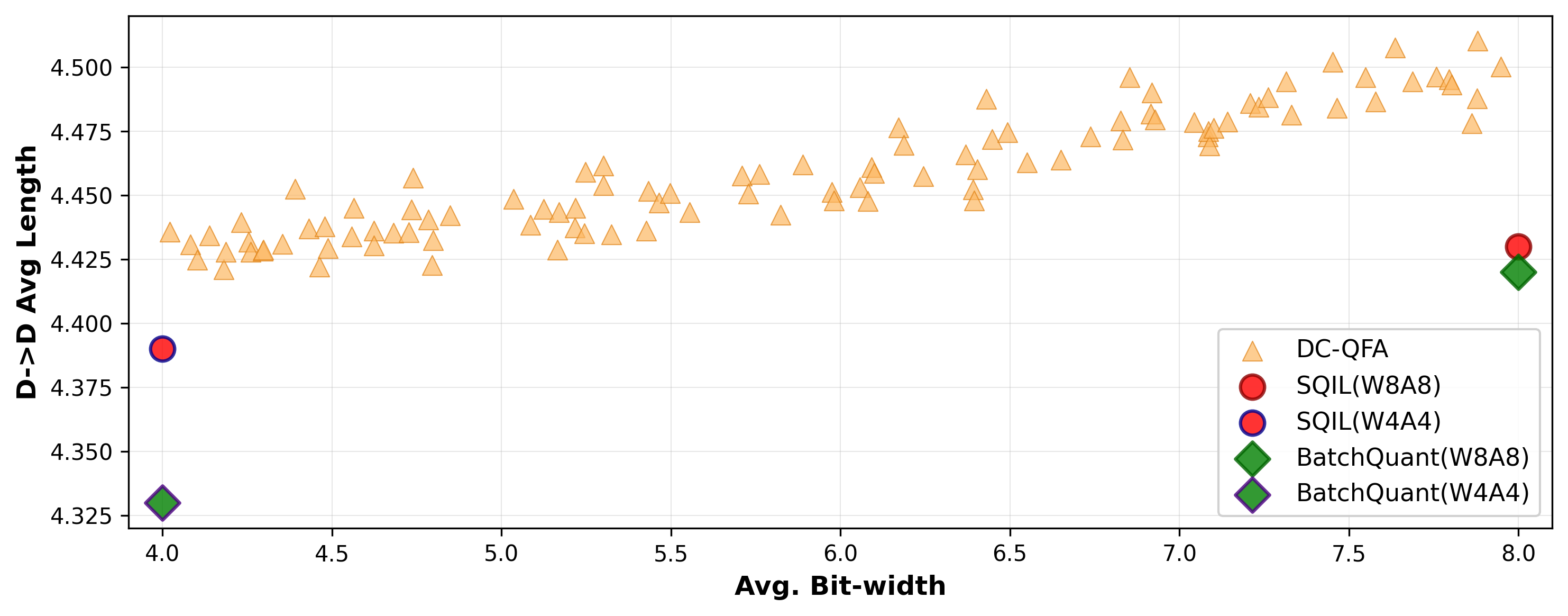}
    \caption{For mixed-precision deployment, the bit-widths $b^W_l$ and $b^A_l$ for each layer are sampled from $\{\text{INT4}, \text{INT8}\}$. DC-QFA achieves superior performance compared to fixed-precision baselines.}
    \label{fig:mixed_precision}
\vspace{-1em}
\end{wrapfigure}
To further demonstrate the capability of DC-QFA in discovering mixed-precision policies, we additionally evaluate heterogeneous bit-width configurations in \cref{fig:mixed_precision}. In this experiment, the average bit-width of subnetworks varies between 4 and 8 bits while individual layers are assigned either INT4 or INT8 precision. The figure shows that the policies discovered by DC-QFA consistently achieve higher rollout performance across a wide range of average bit-widths compared with fixed-precision baselines such as SQIL and BatchQuant. 

\begin{wraptable}{r}{0.45\columnwidth}
\vspace{-2em}
\caption{
Performance, runtime memory, memory reduction, latency, and speedup under different quantization bit-widths on the Jetson Orin NX.
}
  \centering
  \resizebox{0.9\linewidth}{!}{
\begin{tabular}{l c c c}
\toprule
Metric & FP16 & W8A8 & W4A4 \\
\midrule
Memory (GB) & 15.2 & 7.9 & 4 \\
Reduction & -- & 1.92$\times$ & 3.80$\times$ \\
Latency (ms) & 644.74 & 297.84 & 217.38 \\
Speedup & -- & 2.16$\times$ & 2.97$\times$ \\
\bottomrule
\end{tabular}

}
\label{tab:bitwidth_performance}
\vspace{-2em}
\end{wraptable}

\cref{tab:bitwidth_performance} further quantifies the deployment cost of our DC-QFA models on Jetson Orin NX under different bit-widths. The BF16 baseline consumes 15.2GB of running memory and has a per-step latency of 644.74ms. Switching to the INT8 quantization nearly halves the memory footprint to 7.9GB (1.92$\times$ reduction) and reduces latency to 297.84ms, corresponding to a 2.16$\times$ speedup. The INT4 quantization provides the most aggressive compression, bringing memory down to 4GB (3.80$\times$ reduction) and latency to 217.38ms, \ie, almost a 3$\times$ speedup over BF16. Combined with the task performance results, these measurements indicate that our DC-QFA can operate effectively across multiple precision regimes: The INT8 quantization offers a balanced trade-off between efficiency and accuracy, while the INT4 quantization enables maximal resource savings and remains attractive for highly constrained edge deployments.

\section{Conclusion}
\label{sec:conclusion}
We tackle the problem of deploying large transformer-based robotic manipulation policies on heterogeneous hardware with tight resource budgets. Instead of repeating quantization-aware training or pruning for every device, we introduce \textbf{DC-QFA}, a device-conditioned quantization-for-all framework that trains a single mixed-precision supernet and then specializes it to each platform via lightweight NSGA-II search. DC-QFA combines a flexible architecture/bit-width search space, quantization-aware supernet training, and latency/memory regularization from device-specific LUTs, steering optimization toward subnetworks that are both accurate and deployable.

Across three representative policy models, DC-QFA consistently improves the latency--accuracy trade-off over PTQ baselines and matches or surpasses strong quantization and NAS methods. On GPUs and edge devices (\eg, Jetson Orin NX), it delivers $2$--$3\times$ speedups and $2$--$4\times$ reductions in model size and runtime memory with negligible performance drop. To improve long-horizon stability under low precision, we further introduce multi-step on-policy distillation (OPD), which mitigates quantization error accumulation and consistently improves rollout performance on CALVIN under aggressive quantization. Real-world experiments on an Inovo robot equipped with a Robotiq FT 300-S force/torque sensor further confirm that quantized DC-QFA policies can execute stable, contact-rich manipulation under real force feedback. Future work includes extending DC-QFA to broader embodiments and larger foundation models, as well as tighter integration with hardware-specific compilers.

%
%
\bibliographystyle{splncs04}
\bibliography{main}

\begin{thebibliography}{10}
\providecommand{\url}[1]{\texttt{#1}}
\providecommand{\urlprefix}{URL }
\providecommand{\doi}[1]{https://doi.org/#1}

\bibitem{arachchige2025sail}
Arachchige, N.R., Chen, Z., Jung, W., Shin, W.C., Bansal, R., Barroso, P., He, Y.H., Lin, Y.C., Joffe, B., Kousik, S., et~al.: Sail: Faster-than-demonstration execution of imitation learning policies. In: Conf. Robot. Learn. pp. 721--749. PMLR (2025)

\bibitem{quarot}
Ashkboos, S., Mohtashami, A., Croci, M.L., Li, B., Cameron, P., Jaggi, M., Alistarh, D., Hoefler, T., Hensman, J.: Quarot: Outlier-free 4-bit inference in rotated llms. Adv. Neural Inform. Process. Syst.  \textbf{37},  100213--100240 (2024)

\bibitem{batchquant}
Bai, H., Cao, M., Huang, P., Shan, J.: Batchquant: Quantized-for-all architecture search with robust quantizer. In: Adv. Neural Inform. Process. Syst. vol.~34, pp. 1074--1085 (2021)

\bibitem{belkhale2024rt}
Belkhale, S., Ding, T., Xiao, T., Sermanet, P., Vuong, Q., Tompson, J., Chebotar, Y., Dwibedi, D., Sadigh, D.: Rt-h: Action hierarchies using language. In: Robotics Science and Systems (2024)

\bibitem{intelligence2025pi_05}
Black, K., Brown, N., Darpinian, J., Dhabalia, K., Driess, D., Esmail, A., Equi, M.R., Finn, C., Fusai, N., Galliker, M.Y., Ghosh, D., Groom, L., Hausman, K., ichter, b., Jakubczak, S., Jones, T., Ke, L., LeBlanc, D., Levine, S., Li-Bell, A., Mothukuri, M., Nair, S., Pertsch, K., Ren, A.Z., Shi, L.X., Smith, L., Springenberg, J.T., Stachowicz, K., Tanner, J., Vuong, Q., Walke, H., Walling, A., Wang, H., Yu, L., Zhilinsky, U.: $\pi_{0.5}$: a vision-language-action model with open-world generalization. In: Conf. Robot. Learn. (2025)

\bibitem{black2024pi_0}
Black, K., Brown, N., Driess, D., Esmail, A., Equi, M., Finn, C., Fusai, N., Groom, L., Hausman, K., Ichter, B., et~al.: {$\pi_0$}: A vision-language-action flow model for general robot control. arXiv preprint arXiv:2410.24164  (2024)

\bibitem{black2025realtime}
Black, K., Galliker, M.Y., Levine, S.: Real-time execution of action chunking flow policies. In: NeurIPS (2025), \url{https://openreview.net/forum?id=UkR2zO5uww}

\bibitem{black2025training}
Black, K., Ren, A.Z., Equi, M., Levine, S.: Training-time action conditioning for efficient real-time chunking. arXiv preprint arXiv:2512.05964  (2025)

\bibitem{brohan2023rt}
Brohan, A., Brown, N., Carbajal, J., Chebotar, Y., Dabis, J., Finn, C., Gopalakrishnan, K., Hausman, K., Herzog, A., Hsu, J., et~al.: Rt-1: Robotics transformer for real-world control at scale. Robotics Science and Systems  (2023)

\bibitem{brohan2023can}
Brohan, A., Chebotar, Y., Finn, C., Hausman, K., Herzog, A., Ho, D., Ibarz, J., Irpan, A., Jang, E., Julian, R., et~al.: Do as i can, not as i say: Grounding language in robotic affordances. In: Conf. Robot. Learn. pp. 287--318. PMLR (2023)

\bibitem{cai2019once}
Cai, H., Gan, C., Wang, T., Zhang, Z., Han, S.: Once-for-all: Train one network and specialize it for efficient deployment. In: Int. Conf. Learn. Represent. (2019)

\bibitem{cheang2024gr}
Cheang, C.L., Chen, G., Jing, Y., Kong, T., Li, H., Li, Y., Liu, Y., Wu, H., Xu, J., Yang, Y., et~al.: Gr-2: A generative video-language-action model with web-scale knowledge for robot manipulation. arXiv preprint arXiv:2410.06158  (2024)

\bibitem{cheang2025gr}
Cheang, C., Chen, S., Cui, Z., Hu, Y., Huang, L., Kong, T., Li, H., Li, Y., Liu, Y., Ma, X., et~al.: Gr-3 technical report. arXiv preprint arXiv:2507.15493  (2025)

\bibitem{chi2025diffusion}
Chi, C., Xu, Z., Feng, S., Cousineau, E., Du, Y., Burchfiel, B., Tedrake, R., Song, S.: Diffusion policy: Visuomotor policy learning via action diffusion. Int. J. Robot. Res.  \textbf{44}(10-11),  1684--1704 (2025)

\bibitem{gao2024quantnas}
Gao, T., Guo, L., Zhao, S., Xu, P., Yang, Y., Liu, X., Wang, S., Zhu, S., Zhou, D.: Quantnas: quantization-aware neural architecture search for efficient deployment on mobile device. In: IEEE Conf. Comput. Vis. Pattern Recog. Worksh. pp. 1704--1713 (2024)

\bibitem{kim2025fine}
Kim, M.J., Finn, C., Liang, P.: Fine-tuning vision-language-action models: Optimizing speed and success. arXiv preprint arXiv:2502.19645  (2025)

\bibitem{kim2025openvla}
Kim, M.J., Pertsch, K., Karamcheti, S., Xiao, T., Balakrishna, A., Nair, S., Rafailov, R., Foster, E.P., Sanketi, P.R., Vuong, Q., et~al.: Openvla: An open-source vision-language-action model. In: Conf. Robot. Learn. pp. 2679--2713. PMLR (2025)

\bibitem{lin2024awq}
Lin, J., Tang, J., Tang, H., Yang, S., Chen, W.M., Wang, W.C., Xiao, G., Dang, X., Gan, C., Han, S.: Awq: Activation-aware weight quantization for on-device llm compression and acceleration. In: Proceedings of machine learning and systems. vol.~6, pp. 87--100 (2024)

\bibitem{liu2023libero}
Liu, B., Zhu, Y., Gao, C., Feng, Y., Liu, Q., Zhu, Y., Stone, P.: Libero: Benchmarking knowledge transfer for lifelong robot learning. In: Adv. Neural Inform. Process. Syst. vol.~36, pp. 44776--44791 (2023)

\bibitem{liu2025paretoq}
Liu, Z., Zhao, C., Huang, H., Chen, S., Zhang, J., Zhao, J., Roy, S., Jin, L., Xiong, Y., Shi, Y., et~al.: Paretoq: Scaling laws in extremely low-bit llm quantization. In: Adv. Neural Inform. Process. Syst. (2025)

\bibitem{mees2022calvin}
Mees, O., Hermann, L., Rosete-Beas, E., Burgard, W.: Calvin: A benchmark for language-conditioned policy learning for long-horizon robot manipulation tasks. IEEE Robotics and Automation Letters  \textbf{7}(3),  7327--7334 (2022)

\bibitem{o2024open}
O'Neill, A., Rehman, A., Maddukuri, A., Gupta, A., Padalkar, A., Lee, A., Pooley, A., Gupta, A., Mandlekar, A., Jain, A., et~al.: Open x-embodiment: Robotic learning datasets and rt-x models: Open x-embodiment collaboration 0. In: Int. Conf. Robot. Autom. pp. 6892--6903. IEEE (2024)

\bibitem{oquab2024dinov2}
Oquab, M., Darcet, T., Moutakanni, T., Vo, H., Szafraniec, M., Khalidov, V., Fernandez, P., Haziza, D., Massa, F., El-Nouby, A., et~al.: Dinov2: Learning robust visual features without supervision. Trans. Machine Learning Research J.  (2024)

\bibitem{park2025saliency}
Park, S., Kim, H., Kim, S., Jeon, W., Yang, J., Jeon, B., Oh, Y., Choi, J.: Saliency-aware quantized imitation learning for efficient robotic control. In: Int. Conf. Comput. Vis. pp. 13140--13150 (2025)

\bibitem{reuss2024multimodal}
Reuss, M., Ya{\u{g}}murlu, {\"O}.E., Wenzel, F., Lioutikov, R.: Multimodal diffusion transformer: Learning versatile behavior from multimodal goals. In: Robotics Science and Systems (2024)

\bibitem{shen2021once}
Shen, M., Liang, F., Gong, R., Li, Y., Li, C., Lin, C., Yu, F., Yan, J., Ouyang, W.: Once quantization-aware training: High performance extremely low-bit architecture search. In: Int. Conf. Comput. Vis. pp. 5340--5349 (2021)

\bibitem{shukor2025smolvla}
Shukor, M., Aubakirova, D., Capuano, F., Kooijmans, P., Palma, S., Zouitine, A., Aractingi, M., Pascal, C., Russi, M., Marafioti, A., et~al.: Smolvla: A vision-language-action model for affordable and efficient robotics. arXiv preprint arXiv:2506.01844  (2025)

\bibitem{touvron2023llama}
Touvron, H., Martin, L., Stone, K., Albert, P., Almahairi, A., Babaei, Y., Bashlykov, N., Batra, S., Bhargava, P., Bhosale, S., et~al.: Llama 2: Open foundation and fine-tuned chat models. arXiv preprint arXiv:2307.09288  (2023)

\bibitem{wang2019haq}
Wang, K., Liu, Z., Lin, Y., Lin, J., Han, S.: Haq: Hardware-aware automated quantization with mixed precision. In: IEEE Conf. Comput. Vis. Pattern Recog. pp. 8612--8620 (2019)

\bibitem{wu2023unleashing}
Wu, H., Jing, Y., Cheang, C., Chen, G., Xu, J., Li, X., Liu, M., Li, H., Kong, T.: Unleashing large-scale video generative pre-training for visual robot manipulation. In: Int. Conf. Learn. Represent. (2024)

\bibitem{wu2025device}
Wu, Y., Wang, H., Chen, Z., Pang, J., Xu, D.: On-device diffusion transformer policy for efficient robot manipulation. In: Int. Conf. Comput. Vis. pp. 14073--14083 (2025)

\bibitem{xiao2023smoothquant}
Xiao, G., Lin, J., Seznec, M., Wu, H., Demouth, J., Han, S.: Smoothquant: Accurate and efficient post-training quantization for large language models. In: Int. Conf. Machine Learning. pp. 38087--38099. PMLR (2023)

\bibitem{xue2025reactive}
Xue, H., Ren, J., Chen, W., Zhang, G., Fang, Y., Gu, G., Xu, H., Lu, C.: Reactive diffusion policy: Slow-fast visual-tactile policy learning for contact-rich manipulation. In: Robotics Science and Systems. Robotics: Science and Systems Foundation (2025)

\bibitem{yi2025one}
Yi, K., Xu, Y., Chang, H., Meng, Y., Zhang, T., Li, J.: One quantllm for all: Fine-tuning quantized llms once for efficient deployments. In: Annual Meeting Assoc. Comput. Linguistics. pp. 23057--23066 (2025)

\bibitem{zhai2023sigmoid}
Zhai, X., Mustafa, B., Kolesnikov, A., Beyer, L.: Sigmoid loss for language image pre-training. In: Int. Conf. Comput. Vis. pp. 11975--11986 (2023)

\bibitem{zitkovich2023rt}
Zitkovich, B., Yu, T., Xu, S., Xu, P., Xiao, T., Xia, F., Wu, J., Wohlhart, P., Welker, S., Wahid, A., et~al.: Rt-2: Vision-language-action models transfer web knowledge to robotic control. In: Conf. Robot. Learn. pp. 2165--2183. PMLR (2023)

\end{thebibliography}
\end{document}